\documentclass[conference]{IEEEtran}
\IEEEoverridecommandlockouts

\usepackage{cite}
\usepackage{amsmath,amssymb,amsfonts}
\usepackage{algorithmic}
\usepackage{graphicx}
\usepackage{textcomp}
\def\BibTeX{{\rm B\kern-.05em{\sc i\kern-.025em b}\kern-.08em
    T\kern-.1667em\lower.7ex\hbox{E}\kern-.125emX}}

\usepackage{graphicx}
\usepackage{amsmath}
\usepackage{amssymb}
\usepackage{booktabs}

\usepackage[pagebackref,breaklinks,colorlinks]{hyperref}

\usepackage[capitalize]{cleveref}
\crefname{section}{Sec.}{Secs.}
\Crefname{section}{Section}{Sections}
\Crefname{table}{Table}{Tables}
\crefname{table}{Tab.}{Tabs.}

\usepackage[utf8]{inputenc} 
\usepackage[T1]{fontenc}    
\usepackage{hyperref}       
\usepackage{url}            
\usepackage{booktabs}       
\usepackage{amsfonts}       
\usepackage{nicefrac}       
\usepackage{microtype}      

\usepackage[tableposition=top]{caption}

\usepackage[utf8]{inputenc} 
\usepackage[T1]{fontenc}    
\usepackage{hyperref}       
\usepackage{url}            
\usepackage{booktabs}       
\usepackage{amsfonts}       
\usepackage{nicefrac}       
\usepackage{microtype}      
\usepackage{lipsum}		
\usepackage{graphicx}
\usepackage{doi}
\usepackage{multirow}

\usepackage{amsmath,amssymb,mathtools,amsthm,bm,etoolbox}
\usepackage{algorithm}
\usepackage{wrapfig}
\usepackage{subfig}
\usepackage{enumitem}
\usepackage{bbm}
\usepackage{cancel, ulem}

\usepackage[table,xcdraw]{xcolor}

\usepackage{algorithmic, comment}

\newtheorem{proposition}{Proposition}

\usepackage{multicol}
\usepackage{xcolor}
\usepackage{multirow}
\usepackage{arydshln}

\newtheorem{theorem}{Theorem}
\newtheorem{lemma}{Lemma}

\newcommand{\nn}{\|}

\newcommand{\1}{\mathbbm{1}}
\newcommand{\EE}{\mathop{\mathbb{E}}}
\newcommand{\PP}{\mathop{\mathbb{P}}}
\newcommand{\Var}{\text{Var}}

\newcommand{\DD}{\mathcal{D}}
\newcommand{\LL}{\mathcal{L}}
\newcommand{\NN}{\mathcal{N}}
\newcommand{\UU}{\mathcal{U}}

\newcommand{\Max}{\text{max}}
\newcommand{\diag}{\text{diag}}

\newcommand{\erf}{\text{erf}}

\newcommand{\grad}{\nabla}
\DeclareMathOperator{\Tr}{Tr}

\begin{document}

\title{Diverse Gaussian Noise Consistency Regularization for Robustness and Uncertainty Calibration}

\author{\IEEEauthorblockN{Theodoros Tsiligkaridis}
\IEEEauthorblockA{ \textit{MIT Lincoln Laboratory}\\
\textit{AI Technology Group} \\
Lexington MA, USA \\
ttsili@ll.mit.edu}
\and
\IEEEauthorblockN{Athanasios Tsiligkaridis}
\IEEEauthorblockA{\textit{Boston University} \\
\textit{Department of ECE}\\
Boston, MA, USA \\
atsili@bu.edu}
}


\maketitle

\begin{abstract}
Deep neural networks achieve high prediction accuracy when the train and test distributions coincide. In practice though, various types of corruptions occur which deviate from this setup and cause severe performance degradations. Few methods have been proposed to address generalization in the presence of unforeseen domain shifts. In particular, digital noise corruptions arise commonly in practice during the image acquisition stage and present a significant challenge for current methods. In this paper, we propose a diverse Gaussian noise consistency regularization method for improving robustness of image classifiers under a variety of corruptions while still maintaining high clean accuracy. We derive bounds to motivate and understand the behavior of our Gaussian noise consistency regularization using a local loss landscape analysis. Our approach improves robustness against unforeseen noise corruptions by 4.2-18.4\% over adversarial training and other strong diverse data augmentation baselines across several benchmarks. Furthermore, it improves robustness and uncertainty calibration by 3.7\% and 5.5\%, respectively, against all common corruptions (weather, digital, blur, noise) when combined with state-of-the-art diverse data augmentations. Code is available at \url{https://github.com/TheoT1/DiGN}.
\end{abstract}


\section{Introduction} \label{sec:intro}
Deep neural networks are increasingly being used in computer vision and have achieved state-of-the-art performance on image classification \cite{Krizhevsky:2012,He:2015,Huang:2019}. However, when the test distribution differs from the train distribution, performance can suffer as a result, even for mild image corruptions and transformations \cite{Hendrycks:benchmark:2019}. In fact, models have unrealistic behavior when faced with out-of-distribution inputs that arise from synthetic corruptions \cite{Hendrycks:benchmark:2019}, spatial transformations \cite{Engstrom:2019}, and data collection setups \cite{Torralba:2011,Recht:2019}. 
Designing models that provide robustness to unforeseen deviations from the train distribution is highly desirable. Furthermore, calibrated uncertainty on data characterized by domain shifts is also desirable as users can know when models are likely to make errors, improving human-machine teaming.

The majority of the literature on robustness has focused on adversarial robustness against pixel-level $\ell_p$ norm-bounded perturbations \cite{Madry:2018, Zhang:2019}. However, the $\ell_p$ norm-bounded perturbation model for corruptions does not account for common realistic corruptions, such as rotation, translation, blur, and contrast changes. Semantic perturbations that alter image attributes while maintaining a natural look have been also shown to fool classifiers \cite{Joshi:2019, Xiao:2021}. Such domain shifts often correspond to large $\ell_p$ perturbations that circumvent defenses against imperceptible pixel-level perturbations. Sampling the entire space of possible image variations is not practical to protect models against test time failures, and knowledge of the test distribution is not warranted.

The natural approach to defending against a particular fixed distribution shift is to explicitly incorporate such data into the training process \cite{Kang:2019}. However, this paradigm has drawbacks including over-fitting to one type of corruption \cite{Geirhos:2018}; in \cite{Kang:2019}, it was shown that $\ell_\infty$ robustness provides poor generalization to unforeseen attacks; in \cite{Chun:2019} several expensive methods improve robustness at the cost of lower clean accuracy or degraded uncertainty estimation.
Data augmentation policies have been proposed to increase clean accuracy \cite{Cubuk:2019} based on reinforcement learning but are computationally expensive. Uncertainty calibration can be improved by simple ensembles \cite{Lakshminarayanan:2017} and pre-training \cite{Hendrycks:2019:pretrain}. However, it has been shown that calibration degrades significantly under domain shifts \cite{Ovadia:2019}. Recent work \cite{Hendrycks:2020, Hendrycks:ICCV:2021} proposed diverse data augmentations to improve robustness and uncertainty calibration against common corruptions. However, robustness to high frequency noise remains a challenge. 

Digital noise corruptions occur often in real world scenarios due to imaging sensors and/or other devices hardware imperfections, transmission losses, and environmental conditions \cite{DIP:2018, Awad:2019}. Image denoising methods have been proposed for certain noise distributions, e.g. Gaussian, impulse \cite{Chen:2017, Awad:2019} that rely on a-priori knowledge of noise statistics. Advances in this area have been made \cite{Lehtinen:2018, Cavalrons:2021} that do not place strong assumptions on or require much prior knowledge on noise distribution. However, these works focus on reconstruction quality, and not on robustness and uncertainty calibration under unforeseen noise corruptions. 


We propose a training technique that enforces consistency between clean and noisy data samples at different scales to enhance robustness against unforeseen corruptions. Our approach is based on a Gaussian data augmentation strategy coupled with consistency regularization. Viewed through the lens of local loss landscape regularity, our proposed consistency regularization method induces both a smoothing (curvature minimization) and flattening (gradient norm minimization) effect on the local loss landscape. Experimental results show that our method produces models that improve robustness in the presence of various practical unforeseen noise distribution shifts. Furthermore, the combination of our approach with more diverse data augmentations achieves state-of-the-art robustness and uncertainty calibration against all common corruptions.

\noindent Our contributions are summarized below:
\begin{enumerate}
    \item We introduce a consistency loss based on the Kullback-Leibler (KL) divergence that embeds clean examples similarly to diverse Gaussian noise augmentations, and a corresponding training algorithm ($\mathtt{DiGN}$).
    \item We provide an analysis of our Gaussian noise consistency loss that uncovers an inducing effect of smoothing (minimizing curvature) and flattening (minimize gradient norm) of the local loss landscape. Furthermore, we derive probabilistic stability bounds that explicitly involve the local loss geometry characteristics. We additionally provide numerical evidence for this finding.
    \item We experimentally show our approach outperforms several competitive baselines against various unforeseen noise corruptions on CIFAR-10-C, CIFAR-100-C, and Tiny-ImageNet-C datasets. We also show our approach can be effectively combined with diverse data augmentation methods to improve upon state-of-the-art robustness and uncertainty calibration against all common corruptions (weather, digital, noise, blur).
\end{enumerate}
Code is available at \url{https://github.com/TheoT1/DiGN}.

\section{Background and Related Work} \label{sec:background}
We assume labeled data of the form $(x,y)\sim \DD$ drawn from distribution $\DD$. The labels $y$ correspond to $C$ classes. Neural network function $f_\theta(\cdot)$ maps inputs into logits, and $\theta$ are the model parameters. The softmax layer is used to map logits into class probability scores given by $[p_\theta(x)]_c = e^{f_{\theta,c}(x)}/\sum_{l} e^{f_{\theta,l}(x)}$.

\noindent \textbf{Standard Training.} The standard criterion for training deep neural networks is empirical risk minimization ($\mathtt{Standard}$):
\begin{equation} \label{eq:ERM}
    \min_\theta \EE_{(x,y) \sim \DD} \left[ \LL(f_\theta(x),y) \right]
\end{equation}
where the loss is chosen to be the cross-entropy function $\LL(f_\theta(x),y)=-y^T \log p_\theta(x)$. While training using the criterion (\ref{eq:ERM}) yields high accuracy on clean test sets, network generalization performance to various data shifts may suffer.

\noindent \textbf{Adversarial Training against $\ell_p$ Adversary.} Adversarial training ($\mathtt{AT}$) \cite{Madry:2018} is one of the most effective defenses  against $\ell_p$ norm-bounded perturbations which minimizes the adversarial risk,
\begin{equation} \label{eq:AT}
    \min_\theta \EE_{(x,y) \sim \DD} \left[ \max_{\|\delta \|_p\leq \epsilon} \LL(f_\theta(x+\delta),y) \right]
\end{equation}
During the training process, adversarial attacks are computed at inputs $x$ that solve the inner maximization problem. The inner maximization may be solved iteratively using projected gradient descent (PGD) for norms $p\in \{2,\infty\}$, i.e., $\delta^{(k+1)} = \mathcal{P}_{B_p(\epsilon)} (\delta^{(k)} + \alpha \grad_\delta \LL(x+\delta^{(k)},y))$
where $\mathcal{P}_{B_p(\epsilon)}(z) = \arg \min_{u \in B_p(\epsilon)} \|z-u\|_2^2$ is the orthogonal projection onto the constraint set. Frank-Wolfe optimization of the inner maximization is an alternative that offers increased transparency and geometric connections to loss landscape and attack distortion \cite{Tsiligkaridis:2022}. Another robust training approach that trades-off the natural and robust error using smoothing is $\mathtt{TRADES}$ \cite{Zhang:2019}:
\begin{equation}
    \min_\theta \EE_{(x,y) \sim \DD} \Big[\LL(f_\theta(x),y) + \lambda \max_{\|\delta \|_p\leq \epsilon} D(p_\theta(x) \parallel p_\theta(x+\delta)) \Big] \label{eq:TRADES}
\end{equation}
where $D(\cdot \parallel \cdot)$ denotes the Kullback-Leibler divergence defined as $D(p\parallel q)=\sum_c [p]_c \log ([p]_c/[q]_c)$ and the inner maximization is computed using PGD. Related works on defenses to adversarial perturbations include gradient regularization \cite{Ros:2018}, curvature regularization \cite{CURE:2019}, and local linearity regularization \cite{Qin:2019}. These methods optimize robustness only inside the $\epsilon$-ball of the training distribution which does not capture \textit{perceptible} data shifts such as common image corruptions \cite{Hendrycks:benchmark:2019}. Training with arbitrarily large $\epsilon$ in the pixel-wise robustness formulations fails in practice as image quality degrades and clean accuracy suffers due to inherent tradeoff \cite{Tsipras:2019}.

\noindent \textbf{Random Self-Ensemble Training.} In the context of adversarial robustness, a random self-ensemble ($\mathtt{RSE}$) method has been proposed \cite{Liu:2018} based on noise injection at the input layer and each layer of neural networks and was demonstrated to provide good levels of robustness against white-box attacks. Considering noise at the input layer only, the $\mathtt{RSE}$ training criterion is:
\begin{equation} \label{eq:RSE}
    \min_\theta \EE_{(x,y) \sim \DD} \left[ \EE_{\delta \sim \NN(0,\sigma^2 I)} \LL\left(f_\theta(x+\delta),y\right) \right]
\end{equation}
and predictions are ensembled at inference time as $\hat{y}(x) = \arg\max_c \frac{1}{n} \sum_{i=1}^n  [p_\theta(x + \delta_i)]_c$ 
where $\delta_i \sim \NN(0,\sigma^2 I)$. This increases inference time due to the ensemble, is prone to overfitting at a fixed noise level $\sigma$, and cannot handle certain real-world corruptions well, such as contrast and fog.

\noindent \textbf{Diverse Data Augmentation Training.} A recent data augmentation technique, $\mathtt{AugMix}$ \cite{Hendrycks:2020}, was shown to achieve very strong performance against unforeseen corruptions by enforcing a consistency loss coupled with a diverse data augmentation scheme:
\begin{equation}
\min_\theta \EE_{(x,y) \sim \DD} \Big[\LL\left(f_\theta(x),y\right) + \lambda JS(p_\theta(x);p_\theta(x_{a,1});p_\theta(x_{a,2})) \Big] \label{eq:augmix}
\end{equation}
where $JS(p_1;p_2;p_3) = \frac{1}{3}(D(p_1\parallel p_{\text{mix}})+D(p_2\parallel p_{\text{mix}})+D(p_3 \parallel p_{\text{mix}}))$ is the Jensen-Shannon divergence, $p_{\text{mix}} = \frac{1}{3}(p_1+p_2+p_3)$. The augmentations $x_{a,1},x_{a,2}$ are transformations of $x$ formed by mixing composition chains where each chain performs a random finite number of operations chosen from the set $\mathcal{A}=$ \{\texttt{rotate}, \texttt{posterize}, \texttt{shear-x}, \texttt{shear-y}, \texttt{translate-x}, \texttt{translate-y}, \texttt{solarize}, \texttt{equalize}, \texttt{autocontrast}\}. Using diversity and randomness in choosing these operations and mixing weights at different levels of severity during training, this data augmentation method is empirically shown to significantly improve robustness against unforeseen corruptions in comparison to $\mathtt{CutOut}$ \cite{DeVries:2017}, $\mathtt{MixUp}$ \cite{Zhang:2017,Tokozume:2018}, $\mathtt{CutMix}$ \cite{Yun:2019}, and $\mathtt{AutoAugment}$ \cite{Cubuk:2018} schemes. Fourier-based data augmentations have also been shown to be nearly as effective and can improve performance of $\mathtt{AugMix}$ further by including Fourier operations in the set $\mathcal{A}$ \cite{Soklaski:2021}. A closely related data augmentation strategy, $\mathtt{AugMax}$ that combines diversity and hardness by first randomly sampling multiple augmentation operators and then learning an adversarial mixture of these operators, has been shown to be effective \cite{Wang:2021}.

Another data augmentation technique shown to improve performance against such image corruptions is $\mathtt{DeepAugment}$ \cite{Hendrycks:ICCV:2021}, which distorts images by perturbing internal representations of image-to-image networks. These perturbations include zeroing, negating, convolving, transposing, applying activation functions, etc., and it generates semantically similar images for training. Although Patch Gaussian augmentations have been explored where noise is added to randomly selected patches on images \cite{Lopes:2019}, their method did not explore consistency regularization and lacked theoretical analysis on the loss landscape.

While these diverse data augmentation frameworks offer good performance, they still have difficulty dealing with certain noise and blur-based corruptions which can be further improved upon.

\noindent \textbf{Our Approach.} We show that our Gaussian noise consistency regularization approach outperforms these methods under noise domain shifts. In addition, when it is combined with more diverse data augmentations, it achieves near state-of-the-art performance against unforeseen common corruptions (weather, digital, noise, blur).

\section{Diverse Gaussian Noise Consistency Regularization}
Our primary goal is to design a training method to improve generalization against a variety of noise corruptions while fitting in existing pipelines with minimal changes. We introduce a consistency loss that embeds representations of clean examples, $x$, and noisy examples, $x+\delta$, similarly. To increase resiliency against a variety of  noise distributions, we diversify the perturbation statistics by choosing a random noise level $\sigma$ uniformly in the range $[0,\sigma_{\Max}]$ and then generating the random perturbation $\delta \sim \NN(0,\sigma^2 I)$.

Our proposed training criterion is:
\begin{align}
    &\min_\theta \EE_{(x,y) \sim \DD} \Big[ \LL\left(f_\theta(x),y\right) \nonumber \\
    &+ \lambda \EE_{\sigma \sim \UU([0,\sigma_{\Max}])} \EE_{\delta \sim \NN(0,\sigma^2 I)} D(p_\theta(x)\parallel p_\theta(x+\delta)) \Big]    \label{eq:DiGN}
\end{align}
where $p_\theta(\cdot)$ denotes the class probability scores and $D(\cdot \parallel \cdot)$ denotes the Kullback-Leibler divergence defined in Section \ref{sec:background}. 
The classification loss $\LL\left(f_\theta(x),y\right)$ in (\ref{eq:DiGN}) maximizes accuracy on clean examples, while the consistency regularization terms force clean and noisy examples to have similar output distributions. Here, $\lambda \geq 0$ is a regularization parameter, $\UU$ denotes a uniform distribution, and $\NN(0,\sigma^2 I)$ denotes the normal distribution with standard deviation $\sigma$. 

\begin{figure*}[t!]
    \centering
    \includegraphics[width=0.775\textwidth]{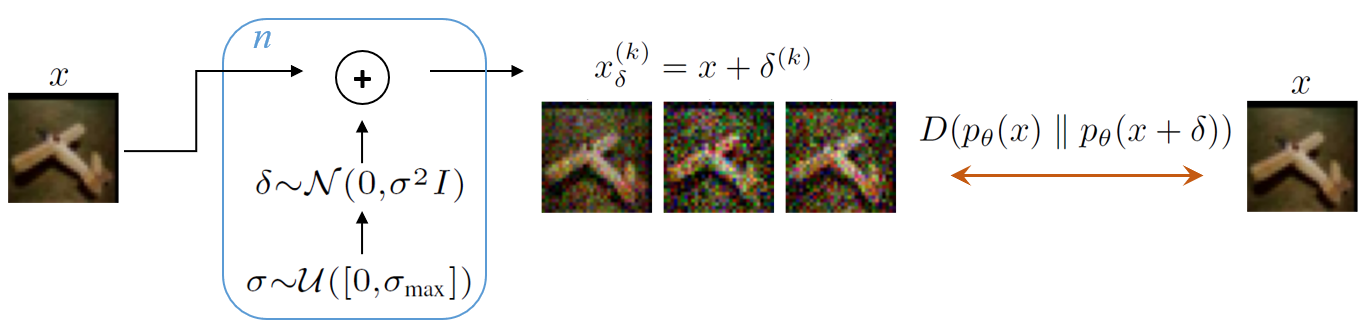}
    \caption{An illustration of our $\mathtt{DiGN}$ training methodology. During training, input images $x$ are fed into $n$ independent branches each associated with a new augmented sample. Each branch adds Gaussian noise $\delta$ to clean images controlled by a variety of scales $\sigma$ sampled from a uniform distribution, and a consistency loss with respect to the clean images is formed. }
    \label{fig:concept}
\end{figure*}

We call our approach (\ref{eq:DiGN}) Diverse Gaussian Noise consistency regularization, $\mathtt{DiGN}$, and Algorithm \ref{alg:DiGN} depicts our training procedure. A sample-based approximation is used to approximate the regularizer to maintain low computational complexity during training. In practice, a couple samples suffice to obtain strong robustness and uncertainty calibration. The concept is illustrated in Figure \ref{fig:concept}.

The motivation for placing distributions on scale parameters $\sigma$ and associated noise perturbations $\delta$ 
to approximate (\ref{eq:DiGN}) is to avoid memorization of fixed scale augmentations \cite{Geirhos:2018}, which leads to poor generalization against various noise domain shifts. In contrast to $\mathtt{RSE}$ (\ref{eq:RSE}), which uses a fixed scale jointly with test-time ensembling at the same scale, we use a diverse set of Gaussian noise scales and do not rely on test-time ensembling. In contrast to $\mathtt{AugMix}$ (\ref{eq:augmix}), $\mathtt{DiGN}$ only makes use of Gaussian noise augmentations and leverages the Kullback-Leibler divergence instead of the Jensen-Shannon divergence as the consistency loss.

\begin{algorithm}[t]
    \caption{ DiGN pseudocode }
    \label{alg:DiGN}
    \begin{algorithmic}
        \STATE{Input:} {Training data $\{(x_i,y_i)\}$, Network $f_\theta$, Training epochs $T$, Batch size $|B|$, learning rate schedule $\eta_t$, hyperparameters $(\lambda,\sigma_{\Max},n)$}
        \STATE{Result:} {Trained network $f_\theta$}
        \FOR{ t=$0$ to $T-1$ }
            \FOR{ each batch $(x,y)\sim \DD$}
                \FOR{ sample $k \in \{1,\dots,n\}$}
                    \STATE{Generate $\sigma^{(k)} \sim \UU(0,\sigma_{\Max})$ for each example}
                    \STATE{Generate $\delta^{(k)} \sim \NN(0,(\sigma^{(k)})^2 I)$ for each example}
                    \STATE{$x_\delta^{(k)} = x+\delta^{(k)}$}
                \ENDFOR
                \STATE{$R(x_i) = \frac{1}{n}\sum_{k=1}^n D(p_\theta(x_i) \parallel p_\theta(x_{i,\delta}^{(k)}))$}
                \STATE{$\LL_T(x_i,y_i) = \LL(x_i,y_i) + \lambda R(x_i)$}
                \STATE{$\theta = \theta - \eta_t \frac{1}{|B|} \sum_{i\in B} \grad_\theta \LL_T(x_i,y_i)$}
            \ENDFOR
        \ENDFOR
    \end{algorithmic}
\end{algorithm}


\subsection{Analysis of Gaussian Noise Consistency Regularization}
In this section, we analyze the effect of Gaussian noise consistency regularization in the small-noise regime. Specifically, we obtain a relationship between loss curvature and Fisher information which is further used to derive a bound on the local loss deviation in a neighborhood of data examples. This bound aids in understanding the effect of Gaussian noise consistency regularization on the local loss landscape. 

Using a second-order Taylor expansion on the KL divergence, for small $\sigma$ \cite{Kullback:1997}: 
\begin{equation*}
    D(p_\theta(x) \parallel p_\theta(x+\delta)) \approx \frac{1}{2} \delta^T G_\theta(x) \delta
\end{equation*}
\vspace{-2.25mm}where $G_\theta(x)$ is the Fisher information matrix (FIM) given by:
\begin{equation} \label{eq:FIM}
    G_\theta(x) = \sum_k [p_\theta(x)]_k \grad_x \log [p_\theta(x)]_k (\grad_x \log [p_\theta(x)]_k)^T
\end{equation}
Taking the expectation,
\begin{equation}
    \EE_{\delta \sim \NN(0,\sigma^2I)} D(p_\theta(x) \parallel p_\theta(x+\delta)) = \frac{\sigma^2}{2} \Tr( G_\theta(x) ) \label{eq:KL_approx0}
\end{equation}
where we used $\EE_{\delta \sim \NN(0,\sigma^2I)}[\delta\delta^T]=\sigma^2 I$. Taking the outer expectation wrt. $\sigma$ on approximation (\ref{eq:KL_approx0}):
\begin{align}
    &\EE_{\sigma \sim \UU([0,\sigma_{\Max}])} \EE_{\delta \sim \NN(0,\sigma^2I)} D(p_\theta(x) \parallel p_\theta(x+\delta)) \nonumber \\
    &\approx \EE_{\sigma \sim \UU([0,\sigma_{\Max}])}\left[\frac{\sigma^2}{2}\right] \Tr( G_\theta(x) ) = \frac{\sigma_{\Max}^2}{6} \Tr( G_\theta(x) ) \label{eq:KL_approx}
\end{align}

The FIM has a strong connection to curvature; in fact for the case of cross-entropy, the Hessian matrix of the loss function is identical to the FIM (\ref{eq:FIM}), as the next proposition shows. 
\begin{proposition} \label{prop:FIM_hessian}
The following relation holds $H(x) := \grad_x^2 \LL(f_\theta(x),y)=G_\theta(x)$.
\end{proposition}
\begin{proof}
The Hessian of the softmax cross-entropy function can be decomposed as (Appendix C in \cite{Qin:2019}) $H(x) = J^T(\diag(p)-pp^T)J$, where $J$ denotes the Jacobian of the network function $f_\theta$ (logits) and $p$ denotes the softmax probabilities. The Jacobian transposed is denoted as $J^T=[J_1^T,\dots,J_K^T]$ with $J_k := \grad_x f_k(x)^T$. Starting from the FIM in (\ref{eq:FIM}), we have:
\begin{align*}
G_\theta(x) &= \sum_k p_k (\grad_x \log p_k)(\grad_x \log p_k)^T \\
    &= \sum_k p_k (\grad_x f_k - J^Tp)(\grad_x f_k - J^Tp)^T \\
    &= \sum_k p_k J_k^T J_k - J^Tpp^TJ \\
    &= J^T (\diag(p) - pp^T) J = H(x)
\end{align*}
\end{proof}

Thus, minimizing with the Gaussian noise regularizer (\ref{eq:KL_approx}) is equivalent to minimizing the curvature in all directions equally since $\Tr(H(x))=\sum_i \lambda_i(H(x))$, where $\lambda_i$ denotes the sorted Hessian eigenvalues. This has the effect of inducing low curvature in the loss landscape of $\LL$ around $x$ and encourages locally linear behavior. 

What we show next is that there is another interesting effect that the KL smoothing regularizer induces. Substituting (\ref{eq:FIM}) into (\ref{eq:KL_approx}) and simplifying, we obtain:
\begin{align}
    \EE_{\sigma \sim \UU([0,\sigma_{\Max}])} &\EE_{\delta \sim \NN(0,\sigma^2)} D(p_\theta(x) \parallel p_\theta(x+\delta)) \nonumber \\
    & \approx \frac{\sigma_{\Max}^2}{6} \sum_k [p_\theta(x)]_k \|\grad_x \log [p_\theta(x)]_k \|_2^2  \label{eq:KL_approx2}
\end{align}
This can be interpreted a regularization term that induces stability of predictions within a local neighborhood of $x$ through weighted logit smoothing. This type of weighted logit smoothing leads to a bound on the local loss deviation.
\begin{theorem} \label{thm:loss_diff}
The following bound holds on the loss function:
\begin{align}
    |\LL(x+\delta,y)&-\LL(x,y)| \leq \| \delta \|_2 \|g(x)\|_2 \nonumber \\
    &+ \frac{1}{2} \lambda_{\text{max}}(H(x)) \|\delta\|_2^2 + o(\| \delta \|_2^2)     \label{eq:theorem_bound} 
\end{align}
where $g(x):=\grad_x \LL(f_\theta(x),y)=\sqrt{ \sum_{k} y_k \| \grad_x \log p(x)_k \|_2^2}$.
\end{theorem}
\begin{proof}
Using the quadratic loss approximation near $x$, we have:
\begin{equation*} \label{eq:loss_quad}
    \LL(x+\delta,y) = \LL(x,y) + \delta^T\grad_x \LL(x,y) + \frac{1}{2}\delta^T H(x)\delta + o(\| \delta \|_2^2)
\end{equation*}
By using up to second order terms, an upper bound on the loss variation holds as:
\begin{align}
    |&\LL(x+\delta,y)-\LL(x,y)| \nonumber \\
        &\approx |\left<\grad_x \LL(x,y), \delta\right> + \frac{1}{2} \delta^T H(x) \delta| \nonumber \\
        &\leq \| \delta \|_2 \cdot \| \grad_x \LL(x,y) \|_2 + \frac{1}{2}\delta^T H(x) \delta  \label{eq:loss_difference}
\end{align}
where we used the triangle inequality, and the Cauchy-Schwarz inequality to bound the linear term and the fact that $H(x)$ is positive semidefinite. The gradient in the first term of (\ref{eq:loss_difference}) can be written as: 
\begin{align*}
    &\| \grad_x \LL(x,y) \|_2 \\
    &= \sqrt{ \| \sum_{k} -y_k \grad_x \log p(x)_k \|_2^2} = \sqrt{ \sum_{k} y_k \| \grad_x \log p(x)_k \|_2^2} \\
    &= \sqrt{ \sum_k e_k \|\grad_x \log p(x)_k\|_2^2 + \sum_k p(x)_k \| \grad_x \log p(x)_k \|_2^2 }
\end{align*}
where $e_k = y_k-p(x)_k$ is the prediction error. The quadratic form in (\ref{eq:loss_difference}) can be upper bounded using the eigenvalue bound $\delta^T H(x)\delta \leq \lambda_{\text{max}}(H(x)) \|\delta\|_2^2$. Using the two preceding bounds into (\ref{eq:loss_difference}), the desired bound (\ref{eq:theorem_bound}) is obtained.
\end{proof}
A consequence of Theorem \ref{thm:loss_diff} is that for correct classifications where $y_k \approx p(x)_k$, minimizing the regularizer (\ref{eq:KL_approx2}) has a twofold effect: (a) curvature is minimized, as the Hessian trace upper bounds the maximum Hessian eigenvalue of the loss $\LL(x,y)$, $\lambda_{\text{max}}(H(x)) \leq \Tr(H(x))$, and (b) loss surface flatness is increased by minimizing the norm of the loss gradient, $\|\grad_x \LL(x,y)\|_2$. The joint effect of minimizing curvature and encouraging flatness in a neighborhood of $x$ implies improved stability of the prediction as the loss changes are small with respect to small perturbations $\delta$ and as a result predictions are invariant to such perturbations.

Next we derive a probabilistic tail bound on the deviation from the true class probability under a Gaussian perturbation model that explicitly involves the gradient and Hessian of the loss function.
\begin{theorem} \label{thm:probabilistic_bound}
Let $\delta$ be a random vector drawn from the Gaussian distribution $\mathcal{N}(0,\sigma^2 I)$. For any $t>0$, the stability bound holds:
\vspace{-1.15mm}
\begin{align*}
    \PP\left(\left|\ln \frac{[p(x)]_y}{[p(x+\delta)]_y}\right|>t\right)
    &\leq C \exp \left( -\frac{t^2}{8d\sigma^2 \nn g(x)\nn_2^2} \right) \\
    & + C \exp \left( -\frac{t}{2d\sigma^2 \lambda_{\max}(H(x))} \right),
\end{align*}

\vspace{-3.3mm}
\noindent where $C=2^d$.
\end{theorem}
\noindent For this proof, we first establish a tail bound on the $\ell_2$ norm of random perturbations $\delta$ in the following lemma.
\begin{lemma} \label{lem:l2_norm_bound}
Consider $\delta$ drawn from the Gaussian distribution $\mathcal{N}(0,\sigma^2 I)$. Then, the following tail bound holds on the norm:
\vspace{-3mm}
\begin{equation} \label{eq:l2_norm_bound}
    \PP(\nn \delta\nn_2 > u) \leq C \exp(-\frac{u^2}{2d\sigma^2}),
\end{equation}

\vspace{-2.53mm}
\noindent where $C=2^d$.
\end{lemma}
\begin{proof}
Using Markov's inequality, we obtain for $s>0$:
\vspace{-2mm}
\begin{align}
\PP(\nn \delta\nn_2>u) &\leq \PP(\nn \delta\nn_1>u)  \leq e^{-su} \EE[e^{s\nn \delta\nn_1}] \nonumber \\
    &= e^{-su} \prod_{i=1}^d \EE[e^{s\sigma |z_i|}] = e^{-su} ( \EE[e^{s\sigma |z_1|}] )^d \label{eq:mgf}
\end{align}

\vspace{-3.25mm}
where we used the decomposition $\nn \delta\nn_1 \equiv \sum_{i=1}^d \sigma |z_i|$ for standard normal variables $z_i$. The folded normal random variable $|z_1|$ has the moment generating function $\EE[e^{v |z_1|}] = 2 e^{\sigma^2v^2/2} \Phi(\sigma v) = e^{\sigma^2v^2/2} (1+\erf(\frac{\sigma v}{\sqrt{2}}))$.
Using this result in (\ref{eq:mgf}), we obtain:
\begin{align*}
     &\PP(\nn \delta\nn_2>u) \leq e^{-su} ( \EE[e^{s\sigma |z_1|}] )^d \\
    &\leq e^{-su} ( e^{\sigma^2 s^2 u^2/2} (1+\erf(\frac{\sigma s u}{\sqrt{2}})) )^d \leq 2^d e^{-su} e^{d \sigma^2 s^2 u^2/2}
\end{align*}
Optimizing the rate by choosing $s=u/(d\sigma^2)$, the bound simplifies to (\ref{eq:l2_norm_bound}). This concludes the proof.
\end{proof}

\noindent Now, we can proceed with the proof of Theorem \ref{thm:probabilistic_bound}. \vspace{-1mm}
\begin{proof}
We have for any $t>0$:
\begin{align*}
    \PP&(|\ln \frac{[p(x)]_y}{[p(x+\delta)]_y}|>t) = \PP(|\LL(x,y)-\LL(x+\delta,y)|>t) \\
        &\stackrel{(a)}{\leq} \PP(\nn \delta\nn_2 \nn g\nn_2 + \frac{1}{2} \lambda_{\max}(H)\nn \delta\nn_2^2 >t) \\
        &\stackrel{(b)}{\leq} \PP(\max\{\nn \delta\nn_2 \nn g\nn_2,  \frac{1}{2} \lambda_{\max}(H)\nn \delta\nn_2^2\}>\frac{t}{2}) 
\end{align*}
\begin{align*}
        &\stackrel{(c)}{\leq} \PP(\{\nn \delta\nn_2 \nn g\nn_2>\frac{t}{2}\}\cup\{\frac{1}{2} \lambda_{\max}(H)\nn \delta\nn_2^2>\frac{t}{2}\}) \\
        &\stackrel{(d)}{\leq} \PP(\nn \delta\nn_2>\frac{t}{2\nn g\nn_2}) + \PP(\nn \delta\nn_2> \sqrt{\frac{t}{\lambda_{\max}(H)}})
\end{align*}
where for (a) we used Theorem \ref{thm:loss_diff}, for (b) $A+B\leq 2 \max\{A,B\}$, for (c) $\{\max\{U,V\}>c\}\subseteq \{U>c\} \cup \{V>c\}$, for (d) we used the union bound. Using the tail bound from Lemma \ref{lem:l2_norm_bound} twice concludes the proof.
\end{proof}

The probabilistic bound in Theorem \ref{thm:probabilistic_bound} shows that the probability of the likelihood ratio $|\ln \frac{[p(x)]_y}{[p(x+\delta)]_y}|$ being large is vanishing exponentially fast at a rate proportional to $(d\sigma^2 \nn g(x) \nn_2^2)^{-1}$ and $(d\sigma^2\lambda_{\max}(H(x)))^{-1}$. This stability result implies that the smaller the gradient norm and Hessian norm of the local loss are, the more stable the model is under random Gaussian perturbations. Since it is expected that models trained with $\mathtt{DiGN}$ exhibit loss surfaces with lower curvature and flatter neighborhoods, the true class likelihood score for noisy data $[p(x+\delta)]_y$ will remain close to that of clean data $[p(x)]_y$.

\subsection{Empirical Validation}
\begin{figure}[t!]
    \centering
    \includegraphics[width=0.485\textwidth]{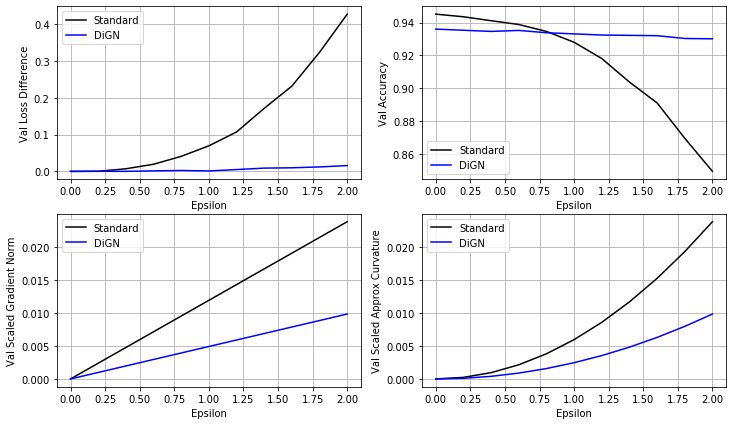}

    \caption{Model resilience against random perturbations $\delta$ of $\ell_2$ radius $\epsilon$ for CIFAR-10 on ResNet-18. Loss difference (top left), validation accuracy (top right), and local loss stability quantified by first (bottom left) and second order (bottom right) information are closely aligned. Our model maintains strong performance across a wide variety of perturbation strengths as opposed to $\mathtt{Standard}$ training, which quickly degrades. }
    \label{fig:loss_acc_curves}
\end{figure}

Figure \ref{fig:loss_acc_curves} (top left) shows the validation loss difference $|\LL(x+\delta,y)-\LL(x,y)|$ to random perturbations $\delta$ of $\ell_2$ radius $\epsilon$ for the CIFAR-10 validation set. These perturbations were generated by first sampling $u\sim \mathcal{N}(0,I)$, and setting $\delta=\epsilon \cdot u/\nn u \nn_2$. Our $\mathtt{DiGN}$ model maintains high validation accuracy (top right), reflected by the loss difference, against perturbations $\delta$ of increasing strength while the $\mathtt{Standard}$ model quickly degrades. The bottom left and right plots show the gradient norm and approximate curvature, which roughly represent the first order term given by $\epsilon \cdot \nn \grad \LL(x,y)\nn_2$, and second order term given by $\frac{1}{2} \epsilon^2 \cdot \Tr(H^TH)$, of the bound (\ref{eq:theorem_bound}). We use a centered difference approximation of the curvature term $\Tr(H^TH)=\EE_{z\sim \mathcal{N}(0,I)} \nn Hz \nn_2^2 \approx \frac{1}{n} \sum_{i=1}^n \frac{1}{2\eta}\nn \grad \LL(x+\eta z_i,y) - \grad \LL(x-\eta z_i,y)\nn_2^2$ where $z_i \sim \mathcal{N}(0,I)$ and $\eta>0$ is small. As expected, our $\mathtt{DiGN}$ model has flatter gradients and lower curvatures than $\mathtt{Standard}$ training. This is a direct consequence of the loss geometry quantified in Theorem \ref{thm:loss_diff} that is induced by diverse Gaussian noise training (\ref{eq:DiGN}).

\section{Experimental Results}

\noindent \textbf{Datasets.} The datasets used for our experimental results are (a) CIFAR-10, (b) CIFAR-100 \cite{cifar_datasets}, and (c) Tiny-ImageNet \cite{tinyimagenet_dataset}. 
Robustness against data shifts is measured by evaluating on CIFAR-10-C, CIFAR-100-C, and Tiny-ImageNet-C \cite{Hendrycks:benchmark:2019}. Each CIFAR corrupted dataset contains a total of $M=18$ corruptions at $J=5$ severity levels. The Tiny-ImageNet corrupted dataset contains $M=14$ corruptions at $J=5$ severity levels. The `gaussian noise' corruption is excluded from our metric evaluations for fairness.


\noindent \textbf{Baselines.} The baseline methods include standard training ($\mathtt{Standard}$), adversarial training ($\mathtt{AT}$), tradeoff-inspired adversarial defense via surrogate-loss minimization ($\mathtt{TRADES}$), random self-ensemble ($\mathtt{RSE}$), diverse augmentation mixing chains ($\mathtt{AugMix}$), deep augmentations ($\mathtt{DeepAugment}$). The adversarial models based on $\mathtt{AT}$ and $\mathtt{TRADES}$ were trained against an $\ell_\infty$ adversary. $\mathtt{DeepAugment}$ was trained using the $\mathtt{Noise2Net}$ approach in \cite{Hendrycks:ICCV:2021}. Training details and hyperparameters can be found in the code.

\noindent \textbf{Performance Metrics.} Classification performance on clean data is measured using test accuracy, i.e., $\frac{1}{n} \sum_{i=1}^n \1_{\{y_i = \hat{y}_i\}}$. Robustness against domain shifts is measured using accuracy on a corrupted validation set for various severity levels. For corruption type $m \in \{1,\dots,M\}$ and severity level $j \in \{1,\dots,J\}$, let $A_{m,j}$ be the corresponding accuracy. The \textit{mean corruption accuracy (mCA)} is defined as: $mCA = \frac{1}{M J} \sum_{m=1}^M \sum_{j=1}^J A_{m,j}$ (\textit{excluding Gaussian noise for fairness}). We denote the mCA score computed over only noise corruptions as \textit{mCA-N} (\textit{excluding Gaussian noise for fairness}).

Classifier calibration scores measure how well the true empirical correct likelihood matches the predicted confidence. Calibration scores provide a signal for establishing trust with human users with applications in safety in high-risk settings, out-of-distribution data detection, and active learning. Uncertainty calibration performance is measured using the root-mean-square  calibration error (\textit{RMSE}), Expected Calibration Error (\textit{ECE}) \cite{Guo:2017, Thulasidasan:2019}, and Overconfidence Error (\textit{OE}) \cite{Thulasidasan:2019}. In addition, we include a Sharpness (SH) metric inspired by \cite{Gneiting:2007}. 
To compute these metrics, we partition the confidence interval $[0,1]$ into $n_B$ bins $\{B_i\}$ and denote the prediction confidence level as $c_j = \max_c [p_\theta(x_j)]_c$. The empirical accuracy and confidence of bin $B_i$ is $\text{acc}(B_i) = \frac{1}{|B_i|} \sum_{j \in B_i} \1_{\{y_j = \hat{y}_j\}}$, and $\text{conf}(B_i) = \frac{1}{|B_i|} \sum_{j\in B_i} c_j$, respectively. \textit{RMSE} measures the discrepancy between the empirical accuracy and the prediction confidence level as: $\textit{RMSE} = \sqrt{\sum_{i=1}^{n_B} \frac{|B_i|}{n} \left( \text{acc}(B_i) - \text{conf}(B_i)  \right)^2 }$. \textit{ECE} is a popular calibration metric that quantifies miscalibration as the absolute difference in expectation between confidence and accuracy as: $\textit{ECE} = \sum_{i=1}^{n_B} \frac{|B_i|}{n} |\text{acc}(B_i) - \text{conf}(B_i)|$. \textit{OE} penalizes predictions by the confidence when the confidence exceeds accuracy, thus leading to high penalties for overconfident bins: $\textit{OE} = \sum_{i=1}^{n_B} \frac{|B_i|}{n} \text{conf}(B_i) \max\{\text{conf}(B_i)-\text{acc}(B_i),0\}$. Sharpness measures precision of confidence intervals, which we define in our context as $\textit{SH} = \left( \sqrt{ \sum_{i=1}^{n_B} \frac{|B_i|}{n} \Var(\{c_j\}_{j \in B_i})} \right)^{-1}$. Intuitively, the larger the SH is, the more concentrated predicted confidence levels are on the reliability plot across all bins. The notation $'\textit{-N}'$ denotes computation only over noise corruptions (excluding Gaussian noise).

\begin{table*}[t]
\centering
\caption{Classification accuracy on clean CIFAR-10/CIFAR-100/Tiny-ImageNet datasets and Mean Corrupted Accuracy (mCA-N) on noise corruption subsets of  CIFAR-10-C/CIFAR-100-C/Tiny-ImageNet-C datasets for ResNet-18 architecture. 
}
\footnotesize
\setlength\tabcolsep{2.5pt} 
\renewcommand{\arraystretch}{0.5} 
\begin{tabular}{cccccccccc}
\hline
\multicolumn{1}{c}{Metric}        & $\mathtt{Standard}$ & $\mathtt{AT}$ & $\mathtt{TRADES}$ & $\mathtt{RSE}$ & $\mathtt{DeepAugment}$ & $\mathtt{AugMix}$ & $\mathtt{AugMax}$ & $\mathtt{DiGN w.o.CR}$ & $\mathtt{DiGN}$     \\ \hline
\multicolumn{10}{c}{\textbf{CIFAR-10}}  \\
\multicolumn{1}{c}{Clean acc.}       & 94.3$\pm$0.1 & 85.8$\pm$0.2 & 83.8$\pm$0.1 & 90.4$\pm$0.1 & 93.0$\pm$0.2 & \textbf{94.9}$\pm$0.2 & 94.1$\pm$0.1 & 91.6$\pm$0.2 & 93.6$\pm$0.1 \\
\multicolumn{1}{c}{mCA-N}            & 60.3$\pm$1.5 & 79.3$\pm$0.8 & 77.8$\pm$0.3 & 85.2$\pm$0.4 & 86.7$\pm$0.3 & 85.0$\pm$0.3 & 82.3$\pm$0.6 & 89.9$\pm$0.2 & \textbf{90.9}$\pm$0.2 \\
\multicolumn{1}{c}{Gaussian Noise} & 48.1$\pm$1.6 & 80.5$\pm$0.8 & 79.0$\pm$0.2 & 87.4$\pm$0.3 & 87.1$\pm$0.4 & 80.2$\pm$0.4 & 75.3$\pm$1.1 & 90.1$\pm$0.2 & \textbf{91.1}$\pm$0.1 \\
\multicolumn{1}{c}{Impulse Noise}    & 57.1$\pm$2.0 & 75.6$\pm$0.7 & 74.1$\pm$0.4 & 80.2$\pm$0.4 & 82.8$\pm$0.2 & 84.6$\pm$0.2 & 82.6$\pm$0.8 & 89.1$\pm$0.1 & \textbf{89.4}$\pm$0.2 \\
\multicolumn{1}{c}{Shot Noise}       & 60.3$\pm$1.2 & 81.4$\pm$0.8 & 79.9$\pm$0.2 & 88.0$\pm$0.3 & 88.7$\pm$0.4 & 84.9$\pm$0.4 & 81.4$\pm$0.5 & 90.3$\pm$0.2 & \textbf{91.7}$\pm$0.2 \\
\multicolumn{1}{c}{Speckle Noise}    & 63.6$\pm$1.2 & 80.9$\pm$0.9 & 79.5$\pm$0.2 & 87.5$\pm$0.4 & 88.7$\pm$0.3 & 85.6$\pm$0.3 & 82.7$\pm$0.4 & 90.4$\pm$0.2 & \textbf{91.7}$\pm$0.1 \\ \hline
\multicolumn{10}{c}{\textbf{CIFAR-100}}                     \\
\multicolumn{1}{c}{Clean acc.}       & 75.8$\pm$0.1 & 59.5$\pm$0.4 & 61.4$\pm$0.2 & 66.7$\pm$0.2 & 71.9$\pm$0.1 & \textbf{76.4}$\pm$0.1 & 75.5$\pm$0.2 & 67.8$\pm$0.2 & 72.8$\pm$0.1 \\
\multicolumn{1}{c}{mCA-N}            & 30.6$\pm$0.6 & 50.0$\pm$1.1 & 50.8$\pm$0.9 & 57.9$\pm$0.7 & 58.5$\pm$0.5 & 57.0$\pm$0.9 & 56.2$\pm$1.0 & 65.5$\pm$0.1 & \textbf{68.4}$\pm$0.1 \\
\multicolumn{1}{c}{Gaussian Noise} & 21.4$\pm$0.6 & 51.1$\pm$0.8 & 53.4$\pm$0.6 & 61.5$\pm$0.9 & 57.2$\pm$0.6 & 47.8$\pm$1.1 & 45.5$\pm$1.0 & 65.3$\pm$0.1 & \textbf{68.6}$\pm$0.0 \\
\multicolumn{1}{c}{Impulse Noise}    & 28.6$\pm$0.5 & 37.8$\pm$1.6 & 43.7$\pm$1.4 & 49.0$\pm$0.7 & 55.1$\pm$0.3 & 60.3$\pm$0.5 & 61.7$\pm$1.2 & 64.0$\pm$0.1 & \textbf{66.0}$\pm$0.2 \\
\multicolumn{1}{c}{Shot Noise}       & 30.8$\pm$0.7 & 52.3$\pm$0.8 & 55.0$\pm$0.6 & 62.9$\pm$0.7 & 60.4$\pm$0.6 & 54.8$\pm$1.0 & 52.7$\pm$0.9 & 66.2$\pm$0.1 & \textbf{69.7}$\pm$0.1 \\
\multicolumn{1}{c}{Speckle Noise}    & 32.3$\pm$0.6 & 50.8$\pm$1.0 & 53.6$\pm$0.7 & 61.7$\pm$0.7 & 59.9$\pm$0.5 & 55.8$\pm$1.1 & 54.3$\pm$1.0 & 66.2$\pm$0.1 & \textbf{69.6}$\pm$0.1 \\ \hline
\multicolumn{10}{c}{\textbf{Tiny-ImageNet}}        \\
\multicolumn{1}{c}{Clean acc.}       & 59.1$\pm$0.4 & 45.5$\pm$0.2 & 46.3$\pm$0.4 & 34.3$\pm$7.8 & 54.9$\pm$0.2 & \textbf{60.4}$\pm$0.1 & 56.7$\pm$4.3 & 51.3$\pm$0.3 & 56.0$\pm$0.2 \\
\multicolumn{1}{c}{mCA-N}            & 24.5$\pm$0.5 & 28.3$\pm$0.1 & 27.7$\pm$0.5 & 23.5$\pm$5.7 & 26.0$\pm$1.2 & 31.3$\pm$1.3 & 30.1$\pm$1.2 & 37.3$\pm$0.2 & \textbf{39.5}$\pm$0.7 \\
\multicolumn{1}{c}{Gaussian Noise} & 22.0$\pm$0.6 & 29.7$\pm$0.1 & 28.6$\pm$0.6 & 23.9$\pm$6.1 & 25.3$\pm$0.9 & 28.8$\pm$1.3 & 27.6$\pm$0.8 & 38.0$\pm$0.2 & \textbf{39.9}$\pm$0.7 \\
\multicolumn{1}{c}{Impulse Noise}    & 23.0$\pm$0.3 & 25.4$\pm$0.1 & 25.2$\pm$0.5 & 23.2$\pm$4.6 & 23.8$\pm$1.5 & 29.8$\pm$1.4 & 28.8$\pm$1.2 & 35.5$\pm$0.1 & \textbf{38.7}$\pm$0.5 \\
\multicolumn{1}{c}{Shot Noise}       & 25.9$\pm$0.7 & 31.1$\pm$0.1 & 30.1$\pm$0.5 & 23.9$\pm$6.7 & 28.2$\pm$0.9 & 32.8$\pm$1.1 & 31.3$\pm$1.1 & 39.0$\pm$0.3 & \textbf{40.2}$\pm$0.8 \\ \hline
\end{tabular}
\label{tab:rob-noise}
\end{table*}

\begin{table*}[t]
\centering
\caption{Uncertainty calibration (RMSE, ECE, OE) and sharpness (SH) metrics for clean and noise-corrupted CIFAR-10/CIFAR-100/Tiny-ImageNet test sets on ResNet-18 architecture.}
\footnotesize
\setlength\tabcolsep{1.5pt}
\renewcommand{\arraystretch}{0.5} 
\begin{tabular}{cccccccccc}
\hline
\multicolumn{1}{c}{Metric}        & $\mathtt{Standard}$ & $\mathtt{AT}$ & $\mathtt{TRADES}$ & $\mathtt{RSE}$ & $\mathtt{DeepAugment}$ & $\mathtt{AugMix}$ & $\mathtt{AugMax}$ & $\mathtt{DiGN w.o.CR}$ & $\mathtt{DiGN}$     \\ \hline
\multicolumn{10}{c}{\textbf{CIFAR-10}}  \\
Clean RMSE     & 5.5$\pm$0.2   & 4.5$\pm$1.3  & 18.7$\pm$0.4 & 6.5$\pm$0.8  & 4.3$\pm$0.2  & 1.5$\pm$0.3  & 10.8$\pm$0.1 & 6.5$\pm$0.2  & 5.2$\pm$0.4  \\
Clean ECE      & 3.3$\pm$0.0   & 2.9$\pm$0.8  & 16.7$\pm$0.4 & 4.4$\pm$0.5  & 2.5$\pm$0.1  & 1.0$\pm$0.2  & 8.7$\pm$0.1  & 4.2$\pm$0.1  & 3.1$\pm$0.1  \\
Clean OE       & 2.9$\pm$0.0   & 0.4$\pm$0.3  & 0.0$\pm$0.0  & 3.8$\pm$0.5  & 2.2$\pm$0.1  & 0.0$\pm$0.0  & 0.0$\pm$0.0  & 3.8$\pm$0.1  & 2.8$\pm$0.1  \\
Clean SH       & 108.3$\pm$0.5 & 61.6$\pm$1.1 & 56.2$\pm$0.3 & 79.9$\pm$1.6 & 85.5$\pm$0.6 & 68.4$\pm$1.4 & 57.8$\pm$0.1 & 85.3$\pm$0.7 & 94.1$\pm$0.9 \\
Corrupt RMSE-N & 30.4$\pm$1.7  & 2.2$\pm$1.0  & 16.7$\pm$0.5 & 9.8$\pm$0.5  & 5.8$\pm$0.2  & 6.5$\pm$0.3  & 3.7$\pm$0.2  & 7.1$\pm$0.2  & 5.6$\pm$0.2  \\
Corrupt ECE-N  & 29.3$\pm$1.9  & 1.7$\pm$0.7  & 14.8$\pm$0.4 & 7.5$\pm$0.5  & 4.2$\pm$0.2  & 4.8$\pm$0.2  & 3.2$\pm$0.2  & 5.0$\pm$0.2  & 3.9$\pm$0.1  \\
Corrupt OE-N   & 26.3$\pm$2.0  & 0.6$\pm$0.4  & 0.0$\pm$0.0  & 6.5$\pm$0.5  & 3.4$\pm$0.2  & 3.9$\pm$0.2  & 0.0$\pm$0.0  & 4.3$\pm$0.1  & 3.4$\pm$0.1  \\
Corrupt SH-N   & 63.8$\pm$0.5  & 58.8$\pm$0.4 & 56.2$\pm$0.1 & 72.5$\pm$1.7 & 67.8$\pm$0.5 & 63.0$\pm$0.6 & 56.7$\pm$0.2 & 80.6$\pm$0.1 & 80.1$\pm$0.3 \\ \hline
\multicolumn{10}{c}{\textbf{CIFAR-100}}  \\
Clean RMSE     & 8.5$\pm$0.1   & 5.2$\pm$2.3  & 4.0$\pm$1.2  & 13.1$\pm$0.4 & 11.2$\pm$0.7 & 2.8$\pm$0.3  & 17.2$\pm$0.5 & 11.9$\pm$0.1 & 6.5$\pm$0.3  \\
Clean ECE      & 7.0$\pm$0.2   & 4.4$\pm$2.2  & 3.2$\pm$1.3  & 11.3$\pm$0.2 & 9.2$\pm$0.5  & 2.1$\pm$0.2  & 15.3$\pm$0.5 & 10.3$\pm$0.0 & 5.4$\pm$0.3  \\
Clean OE       & 5.7$\pm$0.2   & 1.1$\pm$0.8  & 0.0$\pm$0.0  & 8.4$\pm$0.2  & 6.9$\pm$0.4  & 1.5$\pm$0.1  & 0.0$\pm$0.0  & 7.5$\pm$0.1  & 4.1$\pm$0.2  \\
Clean SH       & 61.9$\pm$0.1  & 56.1$\pm$0.1 & 55.5$\pm$0.3 & 58.7$\pm$0.4 & 60.3$\pm$0.5 & 58.8$\pm$0.1 & 55.8$\pm$0.4 & 58.4$\pm$0.3 & 59.3$\pm$0.3 \\
Corrupt RMSE-N & 34.1$\pm$0.8  & 6.9$\pm$2.1  & 2.6$\pm$0.9  & 17.4$\pm$0.6 & 14.6$\pm$0.4 & 17.4$\pm$0.9 & 9.1$\pm$1.3  & 12.3$\pm$0.2 & 9.0$\pm$0.2  \\
Corrupt ECE-N  & 31.5$\pm$0.8  & 6.2$\pm$1.9  & 2.2$\pm$0.7  & 15.6$\pm$0.7 & 13.2$\pm$0.3 & 15.7$\pm$0.9 & 8.3$\pm$1.2  & 10.8$\pm$0.2 & 7.8$\pm$0.2  \\
Corrupt OE-N   & 22.1$\pm$0.8  & 2.6$\pm$1.9  & 0.3$\pm$0.4  & 11.3$\pm$0.5 & 9.2$\pm$0.2  & 11.0$\pm$0.8 & 0.0$\pm$0.0  & 7.7$\pm$0.2  & 5.8$\pm$0.1  \\
Corrupt SH-N   & 55.4$\pm$0.1  & 55.9$\pm$0.1 & 55.7$\pm$0.0 & 57.1$\pm$0.0 & 56.7$\pm$0.1 & 56.6$\pm$0.1 & 56.0$\pm$0.0 & 57.8$\pm$0.2 & 58.4$\pm$0.1 \\
\hline
\multicolumn{10}{c}{\textbf{Tiny-ImageNet}}     \\
Clean RMSE    & 10.3$\pm$0.1 & 8.2$\pm$0.3 & 10.4$\pm$0.3 & 18.0$\pm$0.6 & 7.6$\pm$0.8 & 4.6$\pm$0.6 & 17.8$\pm$1.9 & 10.6$\pm$0.1 & 9.7$\pm$1.1  \\
Clean ECE     & 9.0$\pm$0.2 & 7.4$\pm$0.3  & 9.1$\pm$0.3 & 16.3$\pm$0.4  & 6.5$\pm$0.8  & 3.7$\pm$0.6  & 16.3$\pm$2.0 & 9.5$\pm$0.1  & 8.5$\pm$1.0  \\
Clean OE      & 6.5$\pm$0.0 & 0.0$\pm$0.0  & 0.0$\pm$0.0 & 10.3$\pm$0.3  & 4.0$\pm$0.5  & 1.6$\pm$1.1  & 0.0$\pm$0.0  & 5.8$\pm$0.1  & 5.6$\pm$0.6  \\
Clean SH      & 57.4$\pm$0.4 & 56.4$\pm$0.3  & 56.1$\pm$0.2 & 55.9$\pm$0.1  & 55.7$\pm$0.2  & 56.3$\pm$0.1  & 56.3$\pm$0.1  & 56.0$\pm$0.2  & 56.2$\pm$0.5  \\
Corrupt RMSE-N & 28.9$\pm$0.3 & 2.2$\pm$0.4 & 3.6$\pm$0.2 & 24.1$\pm$0.4 & 19.4$\pm$0.6 & 23.6$\pm$8.3 & 3.6$\pm$2.8 & 16.2$\pm$0.4 & 15.5$\pm$0.8 \\ 
Corrupt ECE-N & 25.7$\pm$0.2 & 2.0$\pm$0.3  & 2.4$\pm$0.1 & 22.3$\pm$0.5  & 19.3$\pm$1.6  & 21.6$\pm$7.9  & 3.1$\pm$2.3  & 14.8$\pm$0.5  & 14.2$\pm$0.8  \\
Corrupt OE-N  & 16.2$\pm$0.3 & 0.0$\pm$0.0  & 0.0$\pm$0.0 & 13.4$\pm$0.4  & 10.4$\pm$1.0  & 13.7$\pm$5.9  & 1.0$\pm$1.4  & 8.6$\pm$0.3  & 8.5$\pm$0.5  \\
Corrupt SH-N  & 55.5$\pm$0.0 & 57.6$\pm$0.2  & 56.9$\pm$0.1 & 55.6$\pm$0.0  & 55.7$\pm$0.1  & 55.5$\pm$0.2  & 56.7$\pm$0.3  & 55.5$\pm$0.1  & 55.5$\pm$0.0  \\
\hline
\end{tabular}
\label{tab:calibr_sharp}
\end{table*}


\noindent \textbf{Robustness Results.} Classification accuracy results computed on the clean and noise-corrupted CIFAR-10, CIFAR-100, and Tiny-ImageNet test sets are shown in Table \ref{tab:rob-noise} for various deep learning methods on the ResNet-18 architecture. For CIFAR-10, our method $\mathtt{DiGN}$ outperforms $\mathtt{Standard}$ baseline by $30.6\%$ mCA-N absolute improvement, and outperforms previous state-of-the-art methods $\mathtt{AugMix}$ and $\mathtt{DeepAugment}$ by large margins of $5.9\%$ and $4.2\%$, adversarial training baselines $\mathtt{AT}$ and $\mathtt{TRADES}$ by $11.6\%$ and $13.1\%$, and $\mathtt{RSE}$ by $5.7\%$. Similarly, for CIFAR-100, $\mathtt{DiGN}$ outperforms $\mathtt{Standard}$ and $\mathtt{AugMix}$ training by $37.8\%$ and $11.4\%$ in mCA-N, respectively. For Tiny-ImageNet, $\mathtt{DiGN}$ outperforms $\mathtt{Standard}$ and $\mathtt{AugMix}$ training by $15\%$ and $8.2\%$ in mCA-N, respectively. Overall, $\mathtt{DiGN}$ consistently achieves the best accuracy against unforeseen digital noise corruptions across all datasets, yielding $4.2-18.4\%$ boost, closing the gap between noise-corrupted and clean accuracy. Hyperparameters were experimentally chosen to achieve a high mCA-N score (see Supplementary Material for sensitivity analysis).

\newlength{\oldintextsep}
\setlength{\oldintextsep}{\intextsep}

Next, we evaluate robustness against all common corruptions (noise, blur, weather, digital). Robustness results on digital noise and all common corruptions are shown in Table \ref{tab:rob_cal-arch} for CIFAR-10/-100 datasets on DenseNet-121 and Wide-ResNet-18 architectures. The Wide-ResNet-18 architecture uses layer widths $320,640,1280,2560$ instead of $64,128,256,512$ used in ResNet-18. Our method $\mathtt{DiGN}$ continues to improve upon prior state-of-the-art $\mathtt{AugMix}$ by $9.3-14.0\%$ in mCA-N. The hybrid combination $\mathtt{DiGN+AugMix}$\footnote{The training loss for this hybrid combination is formed by adding the Jensen-Shannon regularizer from (\ref{eq:augmix}) to the loss (\ref{eq:DiGN}).} outperforms $\mathtt{AugMix}$ by $2.1-3.7\%$ in mCA and $7.0-13.2\%$ in mCA-N. 

The hybrid $\mathtt{DiGN+AugMix}$ consistently achieves the highest mCA and mCA-N scores improving upon previous state-of-the-art $\mathtt{AugMix}$, implying that $\mathtt{DiGN}$ regularizes the model in a complementary way and improves generalization to a variety of digital noise corruptions that $\mathtt{AugMix}$ struggles with as shown in Table \ref{tab:rob-noise}. We believe this is due to $\mathtt{AugMix}$ mostly attending to low-frequency variations, while $\mathtt{DiGN}$ adds a complementary benefit to robustness by providing resilience against high-frequency variations.

\noindent \textbf{Uncertainty Calibration Results.}
Table \ref{tab:calibr_sharp} reports calibration and sharpness metrics for the clean and noise corrupted test sets of CIFAR-10/-100/Tiny-ImageNet on ResNet-18 architecture. 
We remark that while some methods achieve better calibration, their corrupt accuracy lags far behind. For CIFAR-10, $\mathtt{DiGN}$ outperforms $\mathtt{Standard}$ training by $24.8, 25.4, 22.9, 16.3\%$, and outperforms $\mathtt{AugMix}$ by $0.9, 0.9, 0.5, 17.1\%$ in RMSE-N, ECE-N, OE-N and SH-N metrics, respectively. For CIFAR-100, $\mathtt{DiGN}$ outperforms $\mathtt{Standard}$ training by $25.1, 23.7, 16.3, 3.0\%$, and outperforms $\mathtt{AugMix}$ by $8.4, 7.9, 5.2, 1.8\%$ in RMSE-N, ECE-N, OE-N and SH-N metrics, respectively. For Tiny-ImageNet, $\mathtt{DiGN}$ outperforms $\mathtt{Standard}$ training by $13.4, 11.5, 7.7\%$, and outperforms $\mathtt{AugMix}$ by $8.1, 7.4, 5.2\%$ in RMSE-N, ECE-N, OE-N, respectively. Overall, $\mathtt{DiGN}$ offers substantial improvements in RMSE, ECE and OE calibration errors and sharpness across all noise test sets and baseline methods while simultaneously achieving the state-of-the-art in mean corrupt accuracy. Furthermore, the hybrid combination $\mathtt{DiGN+AugMix}$ shown in Table \ref{tab:rob_cal-arch} achieves the lowest RMSE on noise, and all common corruptions for nearly all cases, specifically, $\mathtt{DiGN+AugMix}$ outperforms $\mathtt{AugMix}$ by $1.5-5.5\%$ in RMSE calibration error across all common corruptions.

\noindent \textbf{Ablation Study.} Tables \ref{tab:rob-noise} and \ref{tab:calibr_sharp} show ablation results and demonstrate that consistency regularization (CR) in the $\mathtt{DiGN}$ objective (\ref{eq:DiGN}) outperforms training with diverse Gaussian noise augmentations in the cross-entropy loss (denoted as $\mathtt{DiGN w.o.CR}$).

\begin{table}[h]
\centering
\caption{ Robustness and RMS calibration error for noise and all corruptions on DenseNet-121 and Wide-ResNet-18.
}
\footnotesize
\setlength\tabcolsep{2.5pt} 
\renewcommand{\arraystretch}{0.5} 
\begin{tabular}{ccccc}
\hline
\multicolumn{1}{c}{Metric}            & \multicolumn{1}{c}{$\mathtt{Standard}$} & \multicolumn{1}{c}{$\mathtt{AugMix}$} & \multicolumn{1}{c}{$\mathtt{DiGN}$} & $\mathtt{DiGN}$ \\
\multicolumn{1}{c}{} & \multicolumn{1}{c}{} & \multicolumn{1}{c}{} & \multicolumn{1}{c}{} & $\mathtt{+AugMix}$ \\ \hline
\multicolumn{5}{c}{\textbf{CIFAR-10}}                  \\
\multicolumn{5}{c}{DenseNet-121}                  \\
\multicolumn{1}{c}{Clean acc.}    & 93.9$\pm$0.1  & \textbf{94.6}$\pm$0.1 & 93.0$\pm$0.1 & 94.4$\pm$0.0 \\
\multicolumn{1}{c}{mCA-N}         & 59.5$\pm$1.7  & 82.4$\pm$0.7 & 90.0$\pm$0.2 & \textbf{90.9}$\pm$0.1 \\
\multicolumn{1}{c}{mCA}           & 75.2$\pm$0.3  & 87.9$\pm$0.2 & 85.1$\pm$0.2 & \textbf{90.0}$\pm$0.1 \\
\multicolumn{1}{c}{Clean RMSE}    & 6.3$\pm$0.0  & \textbf{3.3}$\pm$0.1 & 5.9$\pm$0.2 & 4.8$\pm$0.4 \\
\multicolumn{1}{c}{Corrupt RMSE-N}& 32.6$\pm$1.9  & 6.6$\pm$0.2 & 6.5$\pm$0.4 & \textbf{2.5}$\pm$0.4 \\
\multicolumn{1}{c}{Corrupt RMSE}  & 20.7$\pm$0.3  & \textbf{2.5}$\pm$0.0 & 10.6$\pm$0.0 & 2.6$\pm$0.2 \\ \hline
\multicolumn{5}{c}{Wide-ResNet-18}                  \\
\multicolumn{1}{c}{Clean acc.}    & 94.5$\pm$0.2  & 95.5$\pm$0.3 & 94.3$\pm$0.0 & \textbf{95.9}$\pm$0.1 \\
\multicolumn{1}{c}{mCA-N}         & 62.0$\pm$1.6  & 85.7$\pm$0.4 & 91.8$\pm$0.2 & \textbf{92.7}$\pm$0.1 \\
\multicolumn{1}{c}{mCA}           & 78.9$\pm$0.6  & 90.0$\pm$0.2 & 87.7$\pm$0.3 & \textbf{92.1}$\pm$0.1 \\
\multicolumn{1}{c}{Clean RMSE}    & 4.91$\pm$0.0  & \textbf{1.9}$\pm$0.2 & 4.3$\pm$0.5 & 2.0$\pm$0.2 \\
\multicolumn{1}{c}{Corrupt RMSE-N}& 29.0$\pm$1.2  & 8.7$\pm$0.5 & 5.2$\pm$0.2 & \textbf{3.4}$\pm$0.3 \\
\multicolumn{1}{c}{Corrupt RMSE}  & 16.1$\pm$0.5  & 5.1$\pm$0.5 & 8.1$\pm$0.5 & \textbf{3.6}$\pm$0.5 \\ \hline
\multicolumn{5}{c}{\textbf{CIFAR-100}}                  \\
\multicolumn{5}{c}{DenseNet-121}                  \\
\multicolumn{1}{c}{Clean acc.}    & 74.2$\pm$0.2  & \textbf{76.2}$\pm$0.1 & 71.6$\pm$0.1 & 75.6$\pm$0.3 \\
\multicolumn{1}{c}{mCA-N}         & 28.6$\pm$0.6  & 55.4$\pm$0.2 & 66.0$\pm$0.2 & \textbf{68.4}$\pm$0.2 \\
\multicolumn{1}{c}{mCA}           & 47.6$\pm$0.5  & 63.2$\pm$0.2 & 59.2$\pm$0.1 & \textbf{66.6}$\pm$0.3 \\
\multicolumn{1}{c}{Clean RMSE}    & 12.6$\pm$1.9  & \textbf{4.6}$\pm$0.1 & 13.8$\pm$0.1 & 9.4$\pm$0.2 \\
\multicolumn{1}{c}{Corrupt RMSE-N}& 43.5$\pm$2.4  & 7.8$\pm$0.4 & 14.1$\pm$0.3 & \textbf{5.6}$\pm$0.4 \\
\multicolumn{1}{c}{Corrupt RMSE}  & 28.6$\pm$2.6  & \textbf{2.7}$\pm$0.1 & 19.2$\pm$0.3 & 5.5$\pm$0.3 \\ \hline
\multicolumn{5}{c}{Wide-ResNet-18}                  \\
\multicolumn{1}{c}{Clean acc.}    & 75.3$\pm$0.6  & 77.0$\pm$0.5 & 75.3$\pm$0.4 & \textbf{77.6}$\pm$0.2 \\
\multicolumn{1}{c}{mCA-N}         & 39.2$\pm$2.5  & 58.8$\pm$0.5 & 70.5$\pm$0.4 & \textbf{72.0}$\pm$0.2 \\
\multicolumn{1}{c}{mCA}           & 55.1$\pm$0.6  & 65.9$\pm$0.5 & 64.7$\pm$0.5 & \textbf{69.6}$\pm$0.3 \\
\multicolumn{1}{c}{Clean RMSE}    & 5.2$\pm$0.1   & 3.6$\pm$0.5  & 6.1$\pm$0.3  & \textbf{3.3}$\pm$0.6 \\
\multicolumn{1}{c}{Corrupt RMSE-N}& 21.6$\pm$2.2  & 20.8$\pm$0.3 & \textbf{4.1}$\pm$0.0  & 8.9$\pm$0.3 \\
\multicolumn{1}{c}{Corrupt RMSE}  & 11.7$\pm$0.6  & 10.1$\pm$0.3 & 4.7$\pm$0.2  & \textbf{4.6}$\pm$0.2 \\ \hline
\end{tabular}
\label{tab:rob_cal-arch}
\end{table}

\vspace{-2mm}
\section{Conclusion} \vspace{-0.5mm}
A method for training robust deep learning classifiers is presented based on diverse Gaussian noise augmentation at different scales coupled with consistency regularization. It is analytically shown that our approach uncovers smoothing and flattening of the loss landscape. We empirically show that our method yields state-of-the-art robustness and uncertainty calibration under unforeseen noise domain shifts across different network architectures and datasets. In addition, our approach boosts state-of-the-art robustness and uncertainty calibration under unforeseen common corruptions when combined with established diverse data augmentation methods.

\bibliographystyle{IEEEtran}
\bibliography{references}

\end{document}